\documentclass{llncs}

\usepackage[misc]{ifsym}
\usepackage[colorinlistoftodos]{todonotes}
\usepackage{amssymb}

\usepackage[caption=false]{subfig}
\usepackage{float}

\usepackage{multirow}
\usepackage{rotating}

\makeatletter
\def\Hline{%
\noalign{\ifnum0=`}\fi\hrule \@height 1pt \futurelet
\reserved@a\@xhline}
\makeatother

\begin{document}

\title{CNN-based Prostate Zonal Segmentation on T2-weighted MR Images: A Cross-dataset Study}
\titlerunning{MRI Prostate Zonal Segmentation Using CNNs}  
%
\author{Leonardo Rundo\inst{1, 2}(\Letter) \and Changhee Han\inst{3} \and Jin Zhang\inst{3} \and Ryuichiro Hataya\inst{3} \and \\ Yudai Nagano\inst{3} \and Carmelo Militello\inst{2} \and Claudio Ferretti\inst{1} \and Marco S. Nobile \inst{1} \and Andrea Tangherloni \inst{1} \and Maria Carla Gilardi\inst{2} \and Salvatore Vitabile\inst{4}  \and \\ Hideki Nakayama\inst{3} \and Giancarlo Mauri\inst{1}}
\authorrunning{Leonardo Rundo et al.} 
%
%
\institute{Department of Informatics, Systems and Communication,\\University of Milano-Bicocca, Milan, Italy\\
\email{leonardo.rundo@disco.unimib.it}, \and
Institute of Molecular Bioimaging and Physiology (IBFM),\\Italian National Research Council (CNR), Cefal\`{u} (PA), Italy \and
Graduate School of Information Science and Technology,\\The University of Tokyo, Tokyo, Japan \and
Department of Biopathology and Medical Biotechnologies,\\University of Palermo, Palermo, Italy}

\maketitle              

\begin{abstract}
Prostate cancer is the most common cancer among US men. However, prostate imaging is still challenging despite the advances in multi-parametric Magnetic Resonance Imaging (MRI), which provides both morphologic and functional information pertaining to the pathological regions.
Along with whole prostate gland segmentation, distinguishing between the Central Gland (CG) and Peripheral Zone (PZ) can guide towards differential diagnosis, since the frequency and severity of tumors differ in these regions; however, their boundary is often weak and fuzzy. This work presents a preliminary study on Deep Learning to automatically delineate the CG and PZ, aiming at evaluating the generalization ability of Convolutional Neural Networks (CNNs) on two multi-centric MRI prostate datasets.
Especially, we compared three CNN-based architectures: SegNet, U-Net, and pix2pix.
In such a context, the segmentation performances achieved with/without pre-training were compared in $4$-fold cross-validation.
In general, U-Net outperforms the other methods, especially when training and testing are performed on multiple datasets.

\keywords{Prostate zonal segmentation $\cdot$ Prostate cancer $\cdot$ Anatomical MRI $\cdot$ Deep convolutional neural networks $\cdot$ Cross-dataset generalization.}
\end{abstract}
\section{Introduction}
\label{sec:Intro}

Prostate cancer (PCa) is expected to be the most common cancer among US men during 2018~\cite{siegel2018}.
Several imaging modalities can aid PCa diagnosis---such as Transrectal Ultrasound (TRUS), Computed Tomography (CT), and Magnetic Resonance Imaging (MRI)~\cite{rundo2019MedGA}---according to the clinical context. Conventional structural T1-weighted (T1w) and T2-weighted (T2w) MRI sequences can play an important role along with functional MRI, such as Dynamic Contrast Enhanced MRI (DCE-MRI), Diffusion Weighted Imaging (DWI), and Magnetic Resonance Spectroscopic Imaging (MRSI)~\cite{lemaitre2015}.
Therefore, MRI conveys more information for PCa diagnosis than CT, revealing the internal prostatic anatomy, prostatic margins, and the extent of prostatic tumors \cite{villeirs2007}.

The manual delineation of both prostate Whole Gland (WG) and PCa on MR images is a time-consuming and operator-dependent task, which relies on experienced physicians~\cite{rundo2017Inf}. Besides WG segmentation, distinguishing between the Central Gland (CG) and Peripheral Zone (PZ) of the prostate can guide towards differential diagnosis, since the frequency and severity of tumors differ in these regions~\cite{choi2007,niaf2012};
the PZ harbors $70 - 80\%$ of PCa and is a target for prostate biopsy~\cite{haffner2009}. Regarding this anatomic division of the prostate, the zonal compartment scheme proposed by McNeal is widely accepted \cite{selman2011}.
In this context, T2w MRI is the \textit{de facto} standard in the clinical routine of prostate imaging thanks to its high resolution, which allows for differentiating the hyper-intense PZ and hypo-intense CG in young male subjects~\cite{hoeks2011}.
However, the conventional clinical protocol for PCa based on Prostate-Specific Antigen (PSA) and systematic biopsy does not generally obtain reliable diagnostic outcomes; thus, the PZ volume ratio (i.e., the PZ volume divided by the WG volume) was recently integrated for PCa diagnostic refinement~\cite{chang2017}; the CG volume ratio can be also useful for monitoring prostate hyperplasia~\cite{kirby2002}. Furthermore, for robust clinical applications, generalization---among different prostate MRI datasets from multiple institutions---is essential.

So, how can we extract the CG and PZ from the WG on different MRI datasets?
In this work, we automatically segment the CG and PZ using Deep Learning to evaluate the generalization ability of Convolutional Neural Networks (CNNs) on two different MRI prostate datasets.
However, this is challenging since multi-centric datasets are generally characterized by different contrast, visual consistencies, and image characteristics.
Therefore, prostate zones on T2w MR images were manually annotated for supervised learning, and then automatically segmented using a mixed scheme by (\textit{i}) training on either each individual dataset or both datasets and (\textit{ii}) testing on both datasets, using CNN-based architectures: SegNet~\cite{badrinarayanan2017}, U-Net~\cite{ronneberger2015}, and pix2pix~\cite{isola2016}.
In such a context, we compared the segmentation performances achieved with/without pre-training~\cite{tajbakhsh2016}.

The manuscript is structured as follows: Sect. \ref{sec:Background} outlines the state-of-the-art about MRI prostate zonal segmentation methods; Sect. \ref{sec:MatMet} describes the MRI datasets as well as the proposed CNN-based segmentation approach; Sect. \ref{sec:Results} shows our experimental results; finally, some conclusive remarks and possible future developments are given in Sect. \ref{sec:Conclusions}.

\section{Background}
\label{sec:Background}

In prostate MR image analysis, WG segmentation is essential, especially on T2w MR images \cite{ghose2012} or both T2w and the corresponding T1w images \cite{rundo2017Inf,rundo2018WIRN}. Towards it, literature works used atlas-based methods \cite{klein2008}, deformable models, or statistical priors \cite{martin2010}. More recently, Deep Learning \cite{bevilacqua2019} has been successfully applied to this domain, combining deep feature learning with shape models \cite{guo2017} or using CNN-based segmentation approaches \cite{milletari2016}.

Among the studies on prostate segmentation, the most representative works are outlined hereafter. The authors of \cite{toth2013} used active appearance models combined with multiple level sets for simultaneously segmenting prostatic zones. Qiu \textit{et al.} \cite{qiu2014} proposed a zonal segmentation approach introducing a continuous max-flow model based on a convex-relaxed optimization problem with region consistency constraints. Unlike these methods that analyzed T2w images alone, Makni \textit{et al.} \cite{makni2011} exploited the evidential C-means algorithm to partition the voxels into their respective zones by integrating the information conveyed by T2w, DWI, and CE T1w MRI sequences.

However, these methods do not evaluate the generalization ability on different MRI datasets from multiple institutions, making their clinical applicability difficult \cite{albadawy2018}; thus, we verify the cross-dataset generalization using three CNN-based architectures with/without pre-training.
Moreover, differently from a recent CNN-based work on DWI data \cite{clark2017}, to the best of our knowledge, this is the first CNN-based prostate zonal segmentation approach on T2w MRI alone.

\section{Materials and Methods}
\label{sec:MatMet}

For clinical applications with better generalization ability, we evaluate prostate zonal segmentation performances of three CNN-based architectures: SegNet, U-Net, and pix2pix. We also compare the results in $4$-fold cross-validation by ($i$) training on either each individual dataset or both datasets and ($ii$) testing on both datasets, with/without pre-training on a relatively large prostate dataset.

\subsection{MRI Datasets}
\label{sec:Datasets}
We segment the CG and PZ from the WG using two completely different multi-parametric prostate MRI datasets, namely:
\begin{itemize}
\item[$\#1$] dataset containing $21$ patients/$193$ MR slices with prostate, acquired with a whole body Philips Achieva 3T MRI scanner using a phased-array pelvic coil at the Cannizzaro Hospital (Catania, Italy) \cite{rundo2017Inf}.
The MRI parameters: matrix size $= 288 \times 288$ pixels; slice thickness $= 3.0$ mm; inter-slice spacing $=4$ mm; pixel spacing $= 0.625$ mm; number of slices $= 18$;
\item[$\#2$] Initiative for Collaborative Computer Vision Benchmarking (I2CVB) dataset ($19$ patients/$503$ MR slices with prostate), acquired with a whole body Siemens TIM 3T MRI scanner using a body coil at the Hospital Center Regional University of Dijon-Bourgogne (France) \cite{lemaitre2015}.
The MRI parameters: matrix size $\in \{308 \times 384, 336 \times 448, 360 \times 448, 368 \times 448 \}$ pixels; slice thickness $= 1.25$ mm; inter-slice spacing $= 1.0$ mm; pixel spacing $\in \{0.676, 0.721, 0.881,\\ 0.789 \}$ mm; number of slices $= 64$.
\end{itemize}


\begin{figure}[!t]
	\centering
	\subfloat[]{\includegraphics[height=5.0cm]{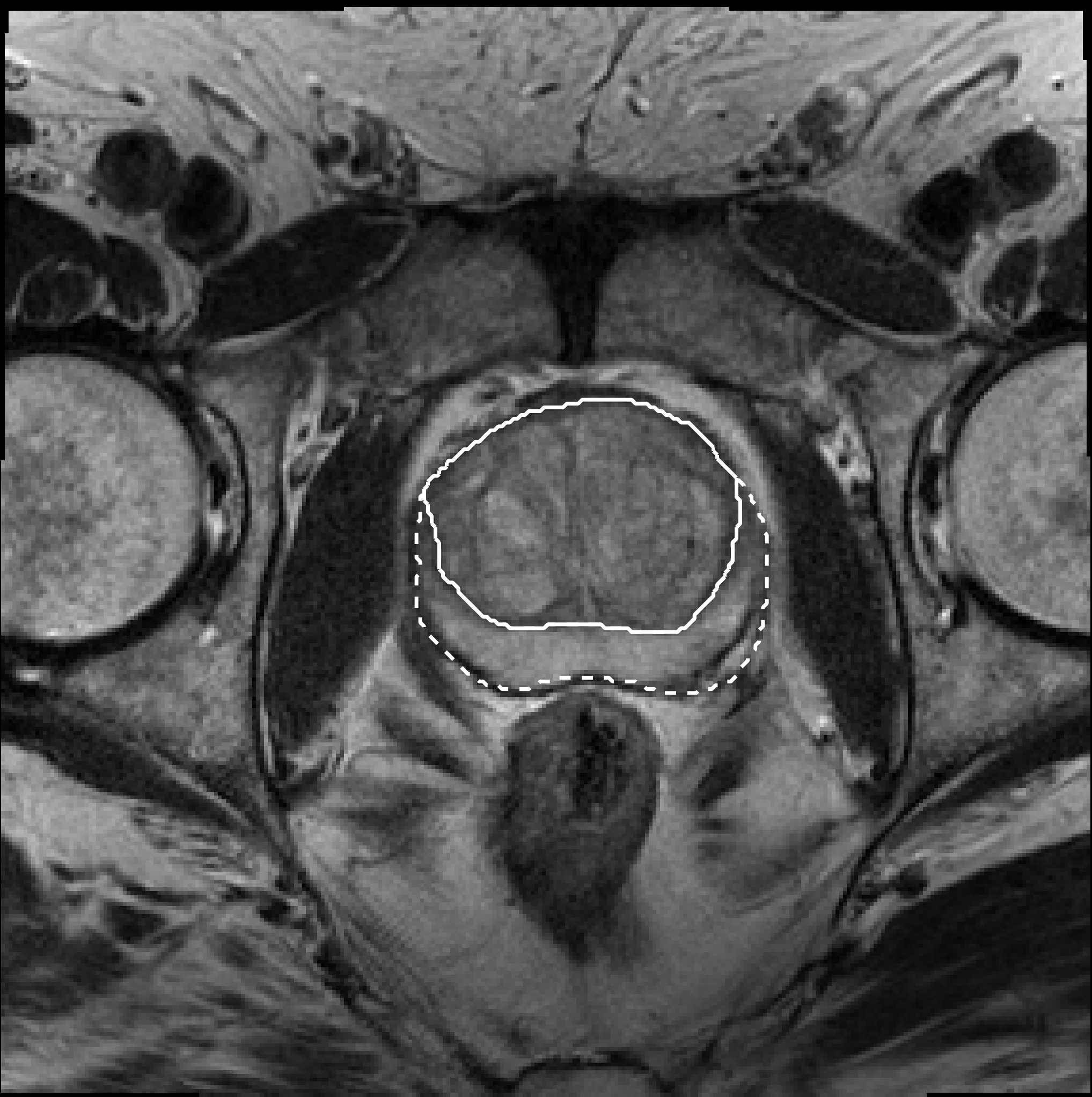}\label{fig:InputImagesA}}\qquad
	\subfloat[]{\includegraphics[height=5.0cm]{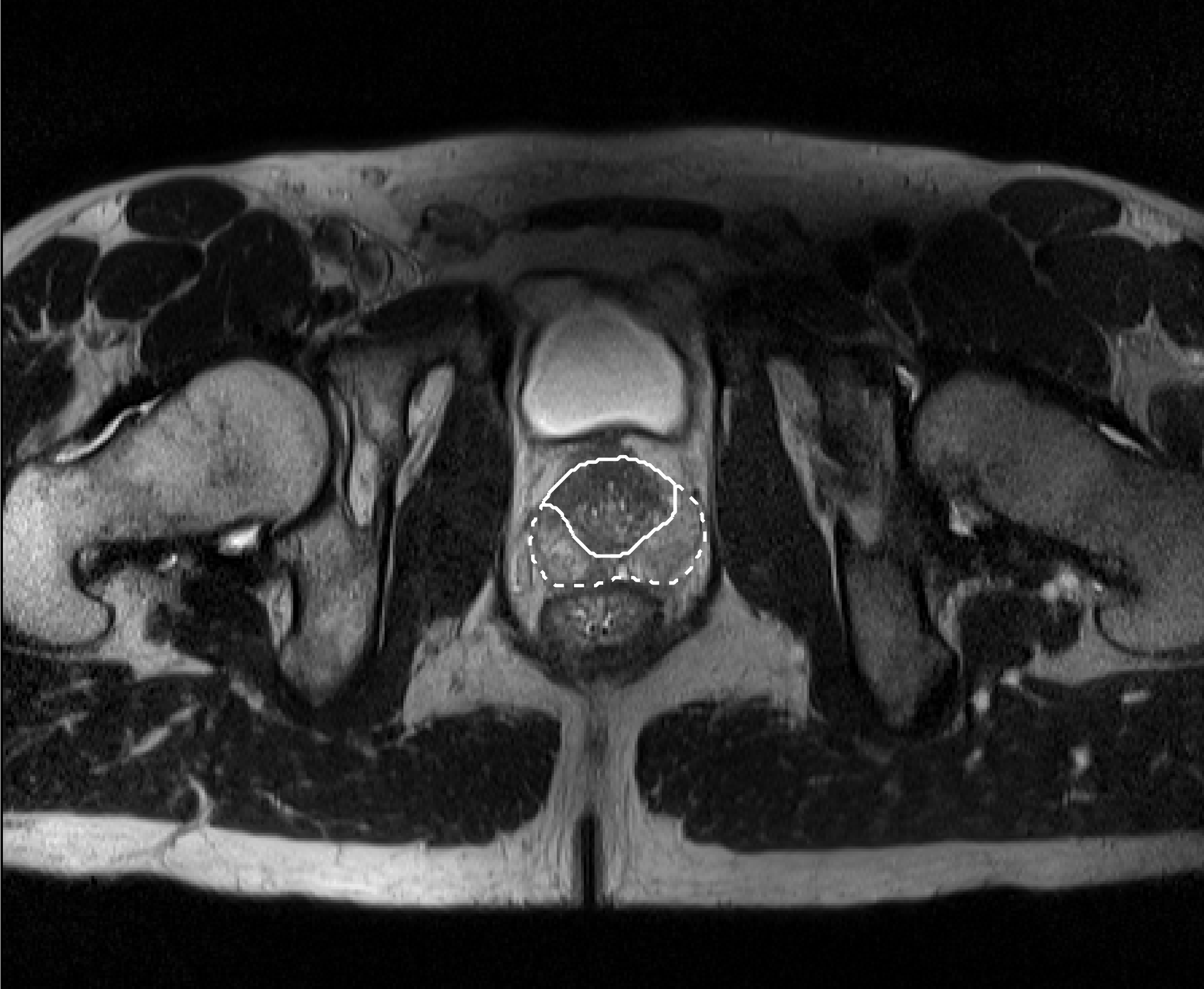}\label{fig:InputImagesB}}\\
	\caption{Example input prostate T2w MR axial slices in their original image ratio: (a) dataset $\#1$; (b)  dataset $\#2$.
    The CG and PZ are highlighted with solid and dashed white lines, respectively.}
	\label{fig:InputImages}	
\end{figure}

To make the proposed approach clinically feasible~\cite{hoeks2011}, we analyzed only T2w images---the most commonly used sequence for prostate zonal segmentation--- among available sequences. Fig. \ref{fig:InputImages} shows two example T2w MR images of the analyzed two datasets.
We conducted the following three experiments using a $4$-fold cross-validation scheme to confirm the generalization effect under different training/testing conditions in our multi-centric study:
\begin{itemize}
\item Individual dataset $\#1$: training on dataset $\#1$ alone, and testing on the whole dataset $\#2$ and the rest of dataset $\#1$ separately for each round;
\item Individual dataset $\#2$: training on dataset $\#2$ alone, and testing on the whole dataset $\#1$ and the rest of dataset $\#2$ separately for each round;
\item Mixed dataset: training on both datasets $\#1$ and $\#2$, and testing on the rest of datasets $\#1$ and $\#2$ separately for each round.
\end{itemize}
For 4-fold cross-validation, we partitioned the datasets $\#1$ and $\#2$ using patient indices $\{ [1, \ldots, 5]$, $[6, \ldots, 10]$, $[11, \ldots, 15]$, $[16, \ldots, 21] \}$ and  $\{ [1, \ldots, 5]$, $[6, \ldots, 10]$, $[11, \ldots, 15]$, $[16, \ldots, 19] \}$, respectively.
Finally, the results from the different cross-validation rounds were averaged.

\subsection{CNN-based Prostate Zonal Segmentation}
\label{sec:Method}

\begin{figure}[!t]
  \includegraphics[width=0.95\textwidth]{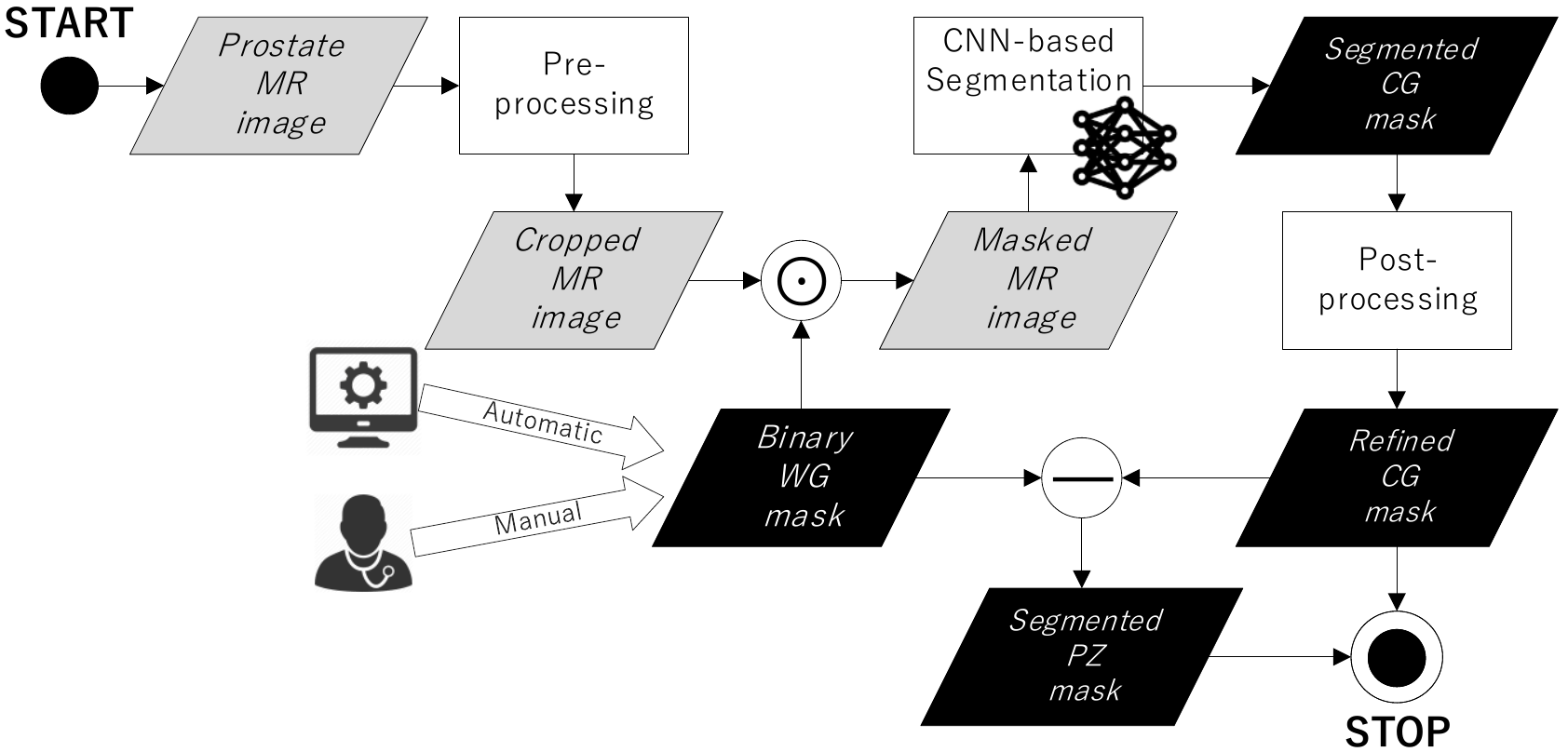}
  \caption{Work-flow of our CNN-based prostate zonal segmentation approach. The gray and black data blocks denote gray-scale images and binary masks, respectively.}
  \label{fig:WorkFlow}
\end{figure}

This work adopts a selective two-step delineation approach to focus on pathological regions in the CG and PZ denoted with $\mathcal{R}_{CG}$ and $\mathcal{R}_{PZ}$, respectively.
Relying on \cite{villeirs2007,qiu2014}, the PZ was obtained by subtracting the CG from the WG ($\mathcal{R}_{WG}$) meeting the constraints: $\mathcal{R}_{WG} = \mathcal{R}_{CG} \cup \mathcal{R}_{PZ}$ and $\mathcal{R}_{CG} \cap \mathcal{R}_{PZ} = \varnothing$.
The overall prostate zonal segmentation method is outlined in Fig. \ref{fig:WorkFlow}.

Starting from the whole prostate MR image, a pre-processing phase, comprising image cropping and resizing to deal with the different characteristics of the two datasets, is performed (see Sect.~\ref{sec:Datasets}).
Afterwards, the resulting image is masked with the binary WG and fed onto the investigated CNN-based models for CG segmentation, which is then refined.
Finally, the PZ delineation is obtained by subtracting the CG from the WG according to~\cite{qiu2014}. 

\subsubsection{Pre-processing}
\label{sec:PreProc}

To fit the image resolution of the dataset $\#1$, we center-cropped the images of the dataset $\#2$ and resized them to $288 \times 288$ pixels.
Furthermore, the images of these datasets were masked using the corresponding prostate binary masks to omit the background and only focus on extracting the CG and PZ from the WG.
This operation can be performed either by an automated method~\cite{rundo2017Inf,rundo2018WIRN} or previously provided manual WG segmentation~\cite{lemaitre2015}. For better training, we randomly cropped the input images from $288 \times 288$ to $256 \times 256$ pixels and horizontally flipped them.

\subsubsection{Investigated CNN-based Architectures}
\label{sec:CNNs}

The following architectures were chosen in our comparative analysis since they cover several aspects regarding the CNN-based segmentation:
SegNet~\cite{badrinarayanan2017} addresses semantic segmentation (i.e., assigning each pixel of the scene to an object class), U-Net~\cite{ronneberger2015} was successfully applied in biomedical image segmentation, while pix2pix~\cite{isola2016} exploits an adversarial generative model to perform image-to-image translations.

During the training of all architectures, the $\mathcal{L}_{DSC}$ loss function (i.e., a continuous version of the Dice Similarity Coefficient) was used  \cite{milletari2016}:
\begin{equation}
	\label{eq:DSCloss}
	\mathcal{L}_{DSC} = - \frac{2\sum_{i=1}^{N} s_i \cdot r_i}{\sum_{i=1}^{N} s_i + \sum_{i=1}^{N} r_i},
\end{equation}
where $s_i$ and $r_i$ represent the continuous values of the prediction map (i.e., the result of the final layer of the CNN) and the ground truth at the $i$-th pixel ($N$ is the total number of pixels to be classified), respectively.

\paragraph{SegNet} is a CNN architecture for semantic pixel-wise segmentation~\cite{badrinarayanan2017}.
More specifically, it was designed for semantic segmentation of road and traffic scenes, wherein classes represent macro-objects, aiming at smooth segmentation results by preserving boundary information.
This non-fully connected architecture, which allows for parameter-efficient implementations suitable for embedded systems, consists of an encoder-decoder network followed by a pixel-wise classification layer.
Since our classification task involves only one class, the soft-max operation and Rectified Linear Unit (ReLU) activation function at the final layer were removed for stable training.

We implemented SegNet using PyTorch.
During the training phase, we used the Stochastic Gradient Descent (SGD)~\cite{bottou2010} with a learning rate of $0.01$, momentum of $0.9$, weight decay of $5\times10^{-4}$, and batch size of $8$.
It was trained for $50$ epochs and the learning rate was multiplied by $0.2$ at the $20$-th and $40$-th epochs.

\paragraph{U-Net} is a fully CNN capable of stable training with a reduced number of samples~\cite{ronneberger2015}, combining pooling operators with up-sampling operations.
The general architecture is an encoder-decoder with skip connections between mirrored layers in the encoder-decoder stacks.
By so doing, high resolution features from the contracting path are combined with the up-sampled output for better localization.
We utilized four scaling operations.
U-Net achieved outstanding performance in biomedical benchmark problems~\cite{falk2019} and has been also serving as an inspiration for novel Deep Learning models for image segmentation.

U-Net was implemented using Keras on top of TensorFlow.
We used SGD with a learning rate of $0.01$, momentum of $0.9$, weight decay of $5 \times 10^{-4}$, and batch size of $4$.
Training was executed for $50$ epochs, multiplying the learning rate by $0.2$ at the $20$-th, and $40$-th epochs.

\paragraph{pix2pix} is an image-to-image translation method coupled with conditional adversarial networks \cite{isola2016}.
As a generator, U-Net is used to translate the original image into the segmented one \cite{ronneberger2015}, preserving the highest level of abstraction.
The generator and discriminator include $8$ and $5$ scaling operations, respectively.

We implemented pix2pix on PyTorch.
Adam~\cite{kingma2014} was used as an optimizer with a learning rate of $0.0002$ and $0.01$ for the discriminator and generator, respectively.
The learning rate for generator was multiplied by $0.1$ every $20$ epochs.
It was trained for $50$ epochs with a batch size of $12$.\\

\vspace{-0.12in}

\subsubsection{Post-processing}
\label{sec:PostProc}

Two simple morphological steps were applied on the obtained CG binary masks to smooth boundaries and avoid disconnected regions:
\begin{itemize}
	\item a hole filling algorithm on the segmented $\mathcal{R}_{CG}$ to remove possible holes in the predicted map;
    \item  a small area removal operation to delete connected components with area less than $\lfloor |\mathcal{R}_{WG}|/8 \rfloor$ pixels, where $|\mathcal{R}_{WG}|$ denotes the number of the pixels contained in the WG segmentation.
This criterion effectively adapts according to the different dimensions of the $\mathcal{R}_{WG}$.
\end{itemize}

\vspace{0.03in}

\subsubsection{Evaluation}
\label{sec:Eval}

The accuracy of the achieved segmentation results $\mathcal{S}$ was quantitatively evaluated with respect to the real measurement (i.e., the gold standard $\mathcal{G}$ obtained manually by experienced radiologists) using the \emph{DSC}:
\begin{equation}
	DSC =  \frac{2|\mathcal{S} \cap \mathcal{G}|} {|\mathcal{S}| + |\mathcal{G}|} \times 
    100\,(\%).
\end{equation}

\subsection{Influence of Pre-training}
\label{sec:PreTraining}

In medical imaging, due to the lack of training data, ensuring CNN's proper training convergence is difficult from scratch.
Therefore, pre-training models on a different application and then fine-tuning is common~\cite{tajbakhsh2016}.

To evaluate cross-dataset generalization abilities \textit{via} pre-training, we compared the performances of the three CNN-based architectures with/without pre-training on a similar application. We used a relatively large dataset of $50$ manually segmented examples from the Prostate MR Image Segmentation 2012 (PROMISE12) challenge \cite{litjens2014}.
Since this competition focuses only on WG segmentation without providing prostate zonal labeling, we pre-trained the architectures on WG segmentation.
To adjust this dataset to our experimental setup, the images of this dataset were resized from $512 \times 512$ to $288 \times 288$ pixels and randomly cropped to $256 \times 256$ pixels; because our task only focuses on slices with prostate, we also omitted initial/final slices without prostate, so the number of slices for each sample was fixed to $25$.

\section{Results}
\label{sec:Results}
This section explains how the three CNN-based architectures segmented the prostate zones, evaluating their cross-dataset generalization ability.

Table \ref{table:results} shows the $4$-fold cross-validation results obtained in the different experimental conditions.
When training and testing are both performed on the dataset $\#1$, U-Net outperforms the other architectures on both CG and PZ segmentation; however, it experiences problems with testing on the dataset $\#2$ due to the limited number of training images in the dataset $\#1$.
In such a case, pix2pix generalizes better thanks to its internal generative model.
When trained on the dataset $\#2$ alone, U-Net yields the most accurate results both in intra- and cross-dataset testing.
This probably derives from the dataset $\#2$'s relatively larger training data as well as U-Net's good generalization ability when sufficient data are available.
Moreover, SegNet reveals rather unstable results, especially when trained on a limited amount of data.
\begin{table*}[t!]
\centering
\begin{footnotesize}
  \caption{Prostate zonal segmentation results of the three CNN-based architectures in $4$-fold cross-validation assessed by the \emph{DSC} (presented as the average and standard deviation). The experimental results are calculated on the different setups of (\textit{i}) training on either each individual dataset or both datasets and (\textit{ii}) testing on both datasets. Numbers in bold indicate the highest \emph{DSC} values for each prostate region (i.e., CG and PZ) among all architectures with/without pre-training (PT).}

\label{table:results}
\begin{tabular}{c|l|c|cc|cc}
\Hline
\multirow{2}{*}{\hspace{30pt}}   & \multicolumn{1}{c|}{\multirow{2}{*}{\textbf{Architecture}}} & \multirow{2}{*}{\textbf{Zone}} & \multicolumn{2}{c|}{\textbf{Testing on Dataset $\#1$}} & \multicolumn{2}{c}{\textbf{Testing on Dataset $\#2$}} \\
                            & \multicolumn{1}{c|}{}                                      &        & \textit{Average}            & \textit{Std. Dev.}            & \textit{Average}          & \textit{Std. Dev.}        \\ \hline

\multirow{12}{*}{\begin{turn}{-90}\textbf{\shortstack{Training on\\Dataset $\#1$}}\end{turn}} & \multirow{2}{*}{SegNet (w/o PT)}                                   & CG     &  80.20                &       3.28              &       74.48           &     5.82             \\
                            &                                                           & PZ     &            80.66        &            11.51          &       59.57           &          12.68         \\

 & \multirow{2}{*}{SegNet (w/ PT)}                                   & CG     &  83.38                &       3.22               &       72.75           &     2.80              \\
                            &                                                           & PZ     &            87.39        &            3.90          &       66.20           &          5.64         \\
\cline{2-7}

                            & \multirow{2}{*}{U-Net (w/o PT)}                                    & CG     &       84.33             &         2.37             &        74.18          &       3.77            \\
                            &                                                           & PZ     &        88.98           &         2.98             &       66.63           &         1.93          \\

 & \multirow{2}{*}{U-Net (w/ PT)}                                   & CG     &  \textbf{86.88}                &       1.60               &       70.11           &     5.31              \\
                            &                                                           & PZ     &            \textbf{90.38}        &            3.38          &       58.89           &          7.06         \\
\cline{2-7}
                            & \multirow{2}{*}{pix2pix (w/o PT)}                                  & CG     &	82.35	&	2.09	&	\textbf{76.61}	&	2.17                 \\
                            &                                                           & PZ     &            87.09	&	2.72	&	73.20	&	2.62                  \\

 & \multirow{2}{*}{pix2pix (w/ PT)}                                   & CG     &  80.38                &       2.81               &       76.19           &     5.77              \\
                            &                                                           & PZ     &            83.53        &            5.65          &       \textbf{73.73}           &          2.40         \\
\hline
\multirow{12}{*}{\begin{turn}{-90}\textbf{\shortstack{Training on\\Dataset $\#2$}}\end{turn}} & \multirow{2}{*}{SegNet (w/o PT)}                                   & CG     &         76.04	&	2.05	&	87.07	&	2.41               \\
                            &                                                           & PZ     &                   77.25	&	3.09	&	82.45	&	1.77                  \\

 & \multirow{2}{*}{SegNet (w/ PT)}                                   & CG     &  77.99                &       2.15               &       87.75           &     2.83              \\

                            &                                                           & PZ     &        76.51	&	2.70	&	82.26	&	2.09                   \\
\cline{2-7}

                            & \multirow{2}{*}{U-Net (w/o PT)}                                    & CG     &      78.88	&	0.88	&	88.21		&	2.10                   \\
                            &                                                           & PZ     &         74.52	&	1.85	&	\textbf{83.03}		&	2.46                  \\
 & \multirow{2}{*}{U-Net (w/ PT)}                                   & CG     &  \textbf{79.82}                &       1.11               &       \textbf{88.66} &     2.28              \\
                            &                                                           & PZ     &        \textbf{74.56}	&	5.12	&	82.48	&	2.47                   \\
\cline{2-7}
                            & \multirow{2}{*}{pix2pix (w/o PT)}                                  & CG     &       77.90	&	0.73	&	86.95	&	2.93                   \\
                            &                                                           & PZ     &                   66.09	&	3.07	&	81.33	&	0.90                   \\
 & \multirow{2}{*}{pix2pix (w/ PT)}                                   & CG     &  77.21                &       1.02               &       85.94           &     4.31              \\
                            &                                                           & PZ     &        67.39	&	5.04	&	80.07	&	0.84                   \\
\hline

\multirow{12}{*}{\begin{turn}{-90}\textbf{\shortstack{Training on\\Mixed Dataset}}\end{turn}} & \multirow{2}{*}{SegNet (w/o PT)}                                   & CG     &          84.28	&	3.12	&	87.92	&	2.80                   \\

                            &                                                           & PZ     &        87.74	&	1.66	&	82.21	&	0.79                   \\

 & \multirow{2}{*}{SegNet (w/ PT)}                                   & CG     &  86.08                &       1.92               &       87.78           &     2.75              \\

                            &                                                           & PZ     &        89.53	&	3.28	&	82.39	&	1.50                   \\
\cline{2-7}
                            & \multirow{2}{*}{U-Net (w/o PT)}                                    & CG     &          \textbf{86.34}	&	2.10	&	\textbf{88.12}		&	2.34                   \\

                            &                                                           & PZ     &           90.74	&	2.40	&	\textbf{83.04}	&	2.30                  \\

 & \multirow{2}{*}{U-Net (w/ PT)}                                   & CG     &  85.82 	               &       1.98               &       87.42           &     1.89              \\

                            &                                                           & PZ     &        \textbf{91.44}	&	2.15	&	82.17	&	2.11                   \\
\cline{2-7}
                            & \multirow{2}{*}{pix2pix (w/o PT)}                                  & CG     &   83.07	&	3.39	&	86.39	&	3.16                  \\

                            &                                                           & PZ     &        83.53	&	2.36	&	80.40	&	1.80                   \\

 & \multirow{2}{*}{pix2pix (w/ PT)}                                   & CG     &  82.08                &       4.37               &       85.96           &     5.40              \\
                            &                                                           & PZ     &        83.04	&	4.20	&	80.60	&	1.49                   \\
\Hline       
\end{tabular}
\end{footnotesize}
\end{table*}
Finally, when trained on the mixed dataset, all three architectures---especially U-Net---achieve good results on both datasets without losing accuracy compared to training on the same dataset alone.
Therefore, using mixed MRI datasets during training can considerably improve the performance in cross-dataset generalization towards other clinical applications.
Comparing the CG and PZ segmentation, when tested on the dataset $\#1$, the results on the PZ are generally more accurate, except when trained on the dataset $\#2$ alone; however, for the dataset $\#2$, segmentations on the CG are generally more accurate.

Fine-tuning after pre-training sometimes leads to slightly better results than training from scratch, when trained only on a single dataset.
However, its influence is generally negligible or rather negative, when trained on the mixed dataset. This modest impact is probably due to the ineffective data size for pre-training.

\begin{figure}[!t]
	\centering
	\subfloat[]{\includegraphics[height=2.5cm]{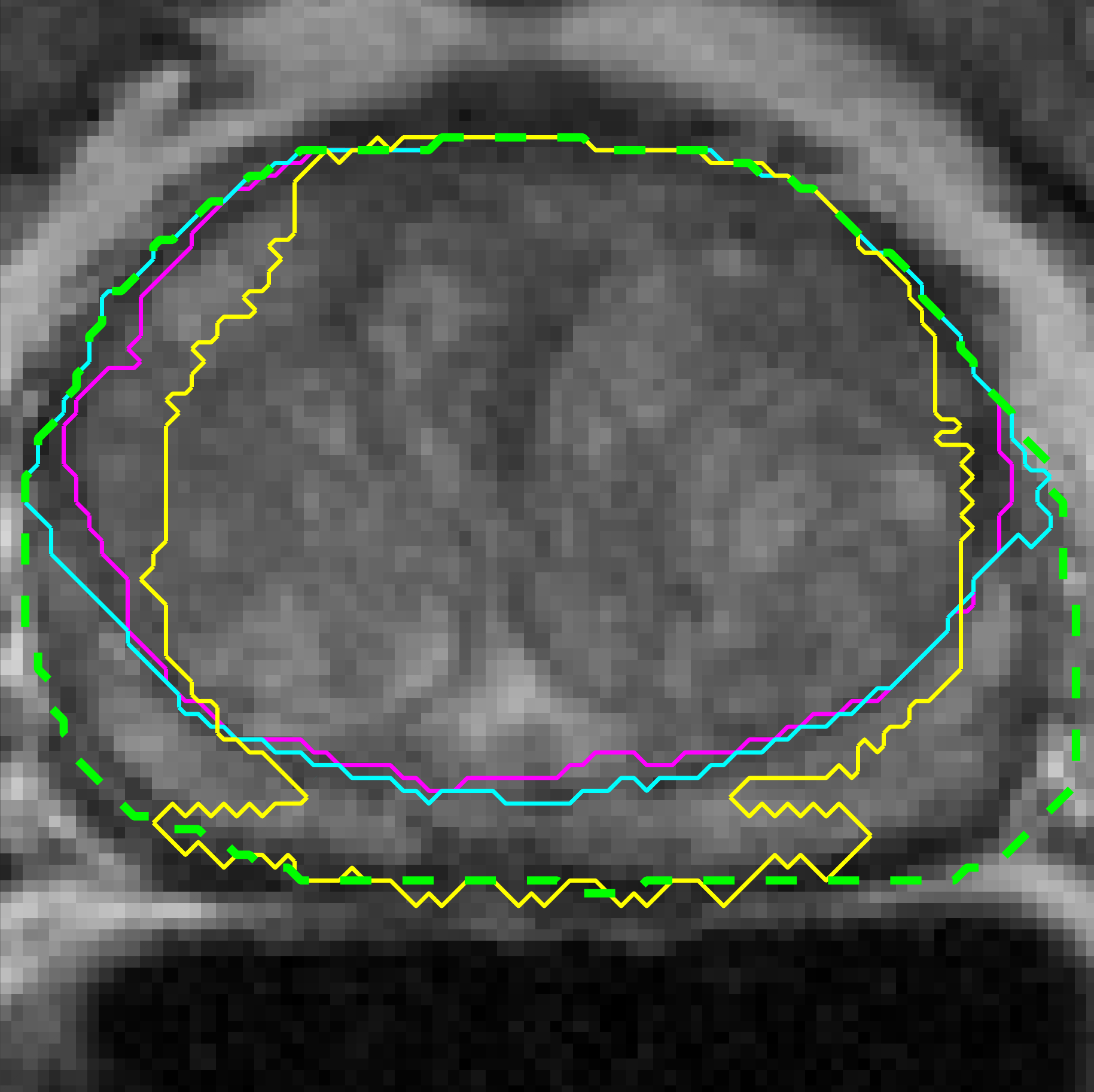}\label{fig:ResultImagesA}}\qquad
\subfloat[]{\includegraphics[height=2.5cm]{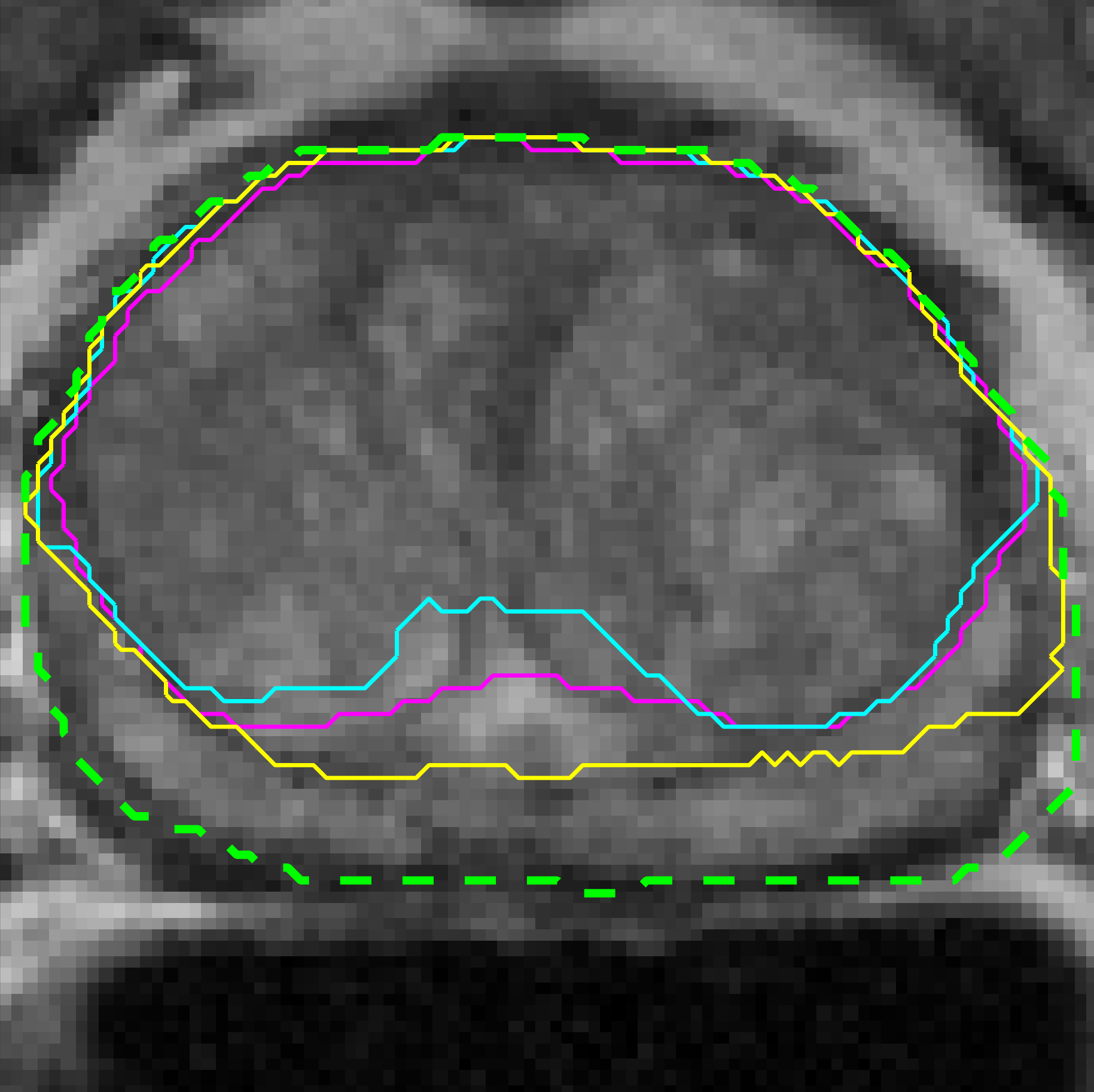}\label{fig:ResultImagesB}}\qquad
\subfloat[]{\includegraphics[height=2.5cm]{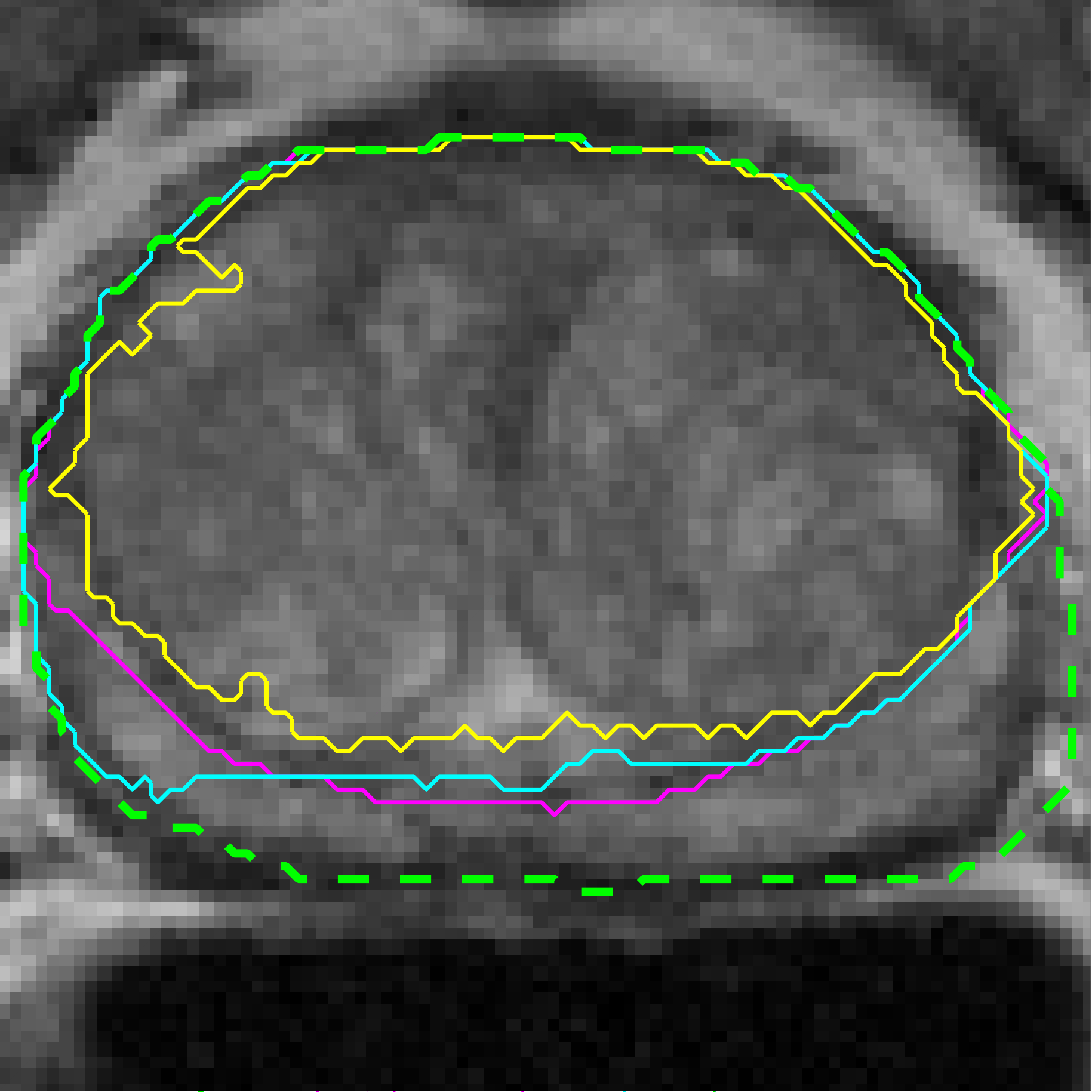}\label{fig:ResultImagesC}}\qquad
\subfloat[]{\includegraphics[height=2.5cm]{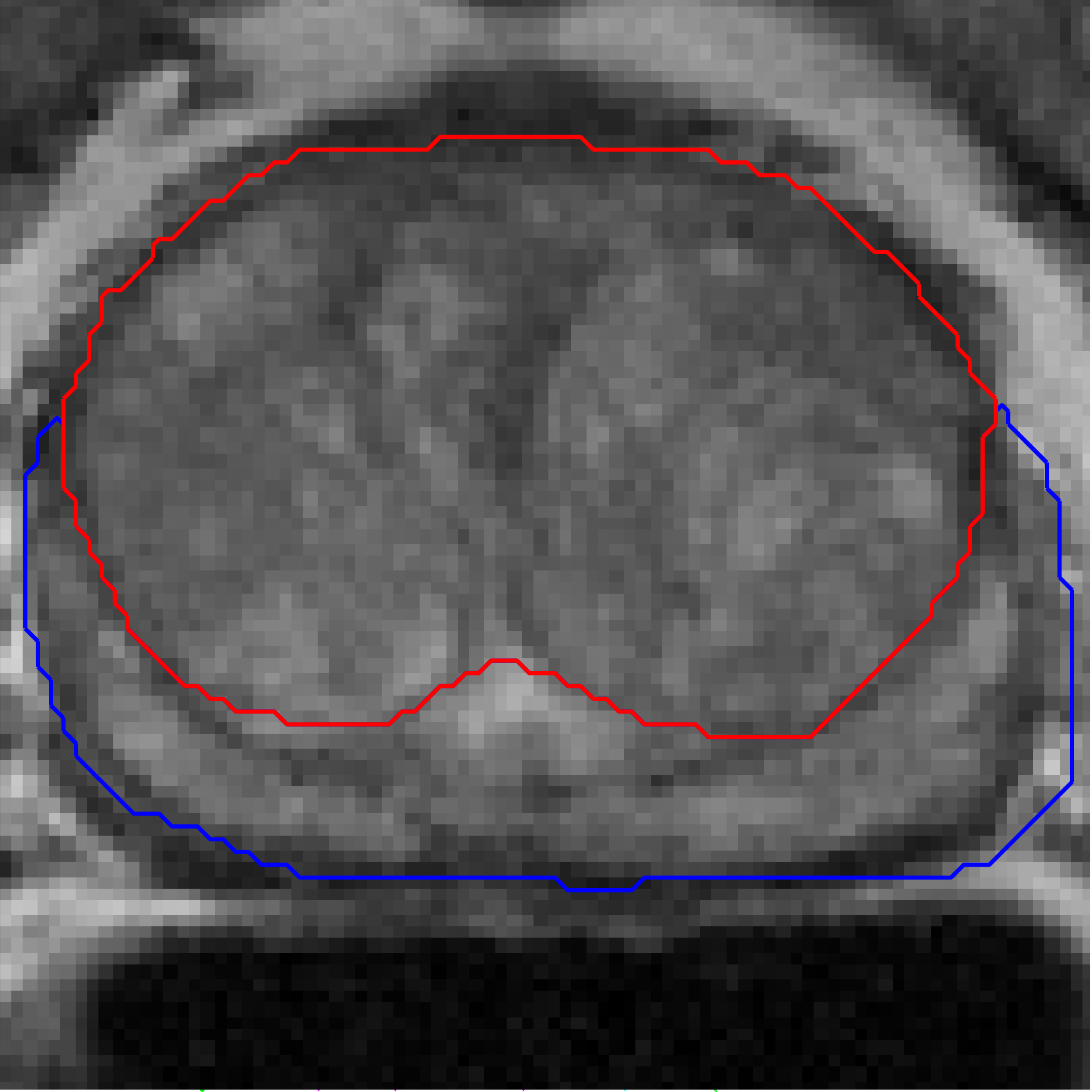}\label{fig:ResultImagesD}} \\
\subfloat[]{\includegraphics[height=2.5cm]{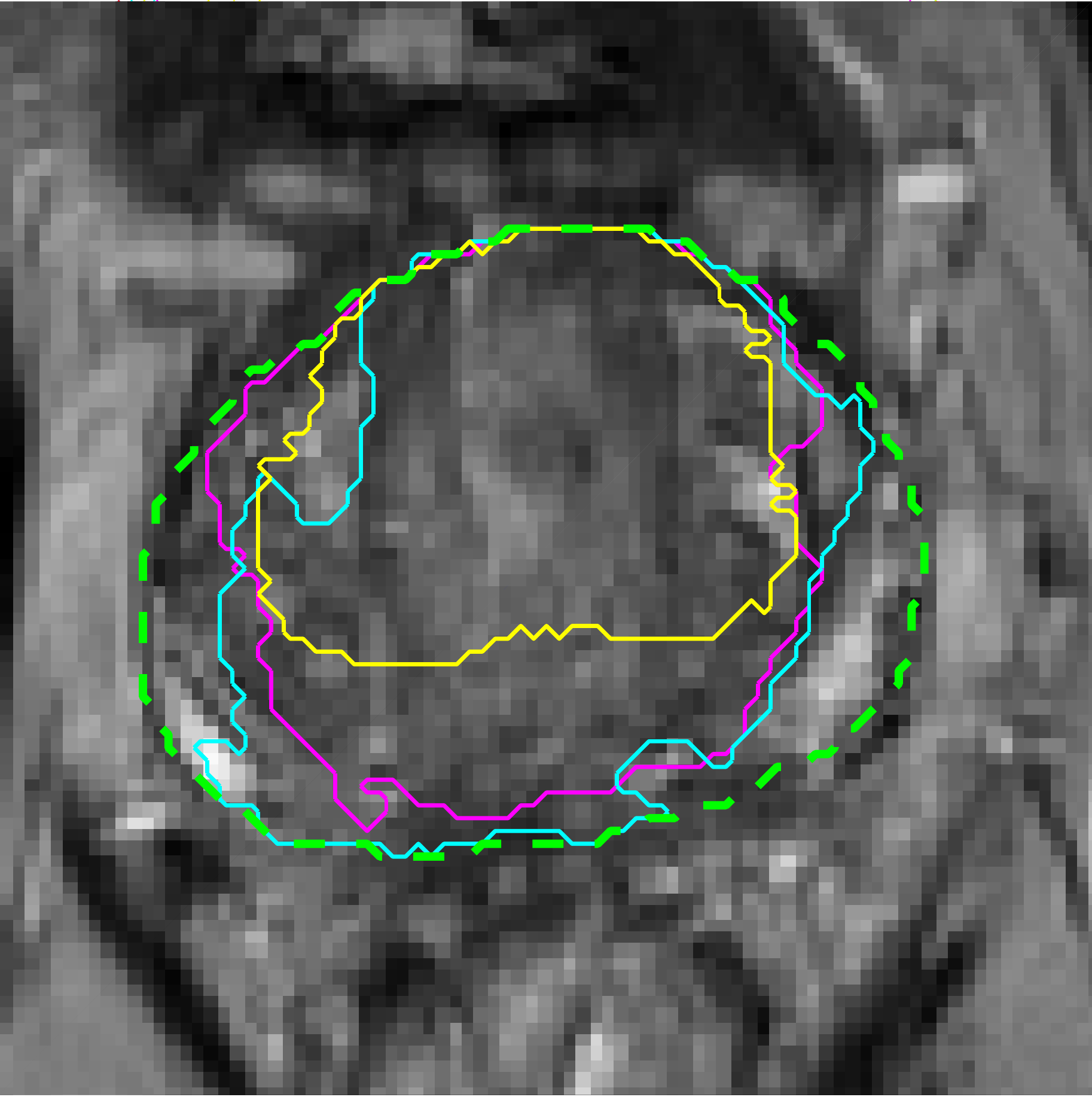}\label{fig:ResultImagesE}}\qquad
\subfloat[]{\includegraphics[height=2.5cm]{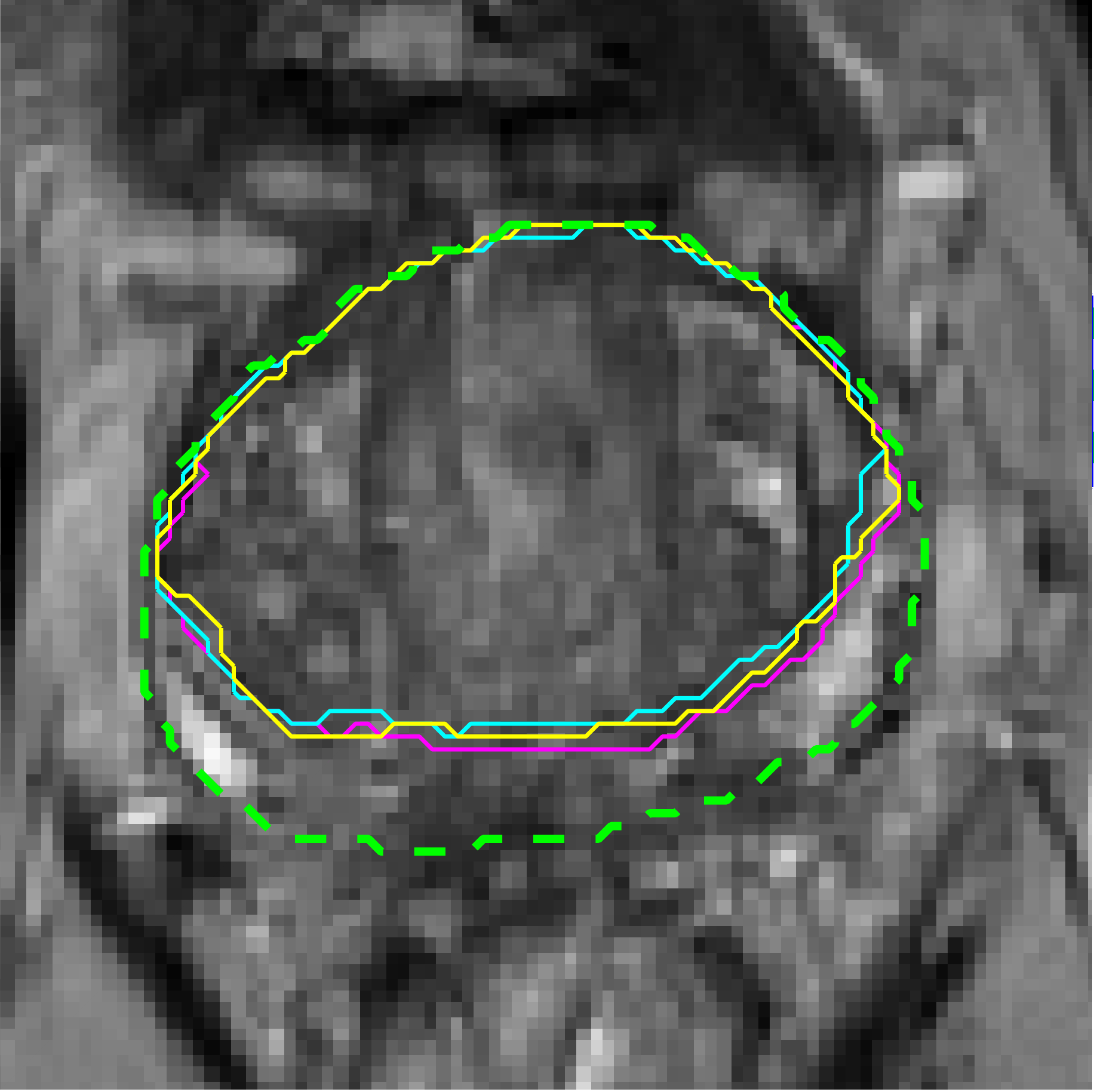}\label{fig:ResultImagesF}}\qquad
\subfloat[]{\includegraphics[height=2.5cm]{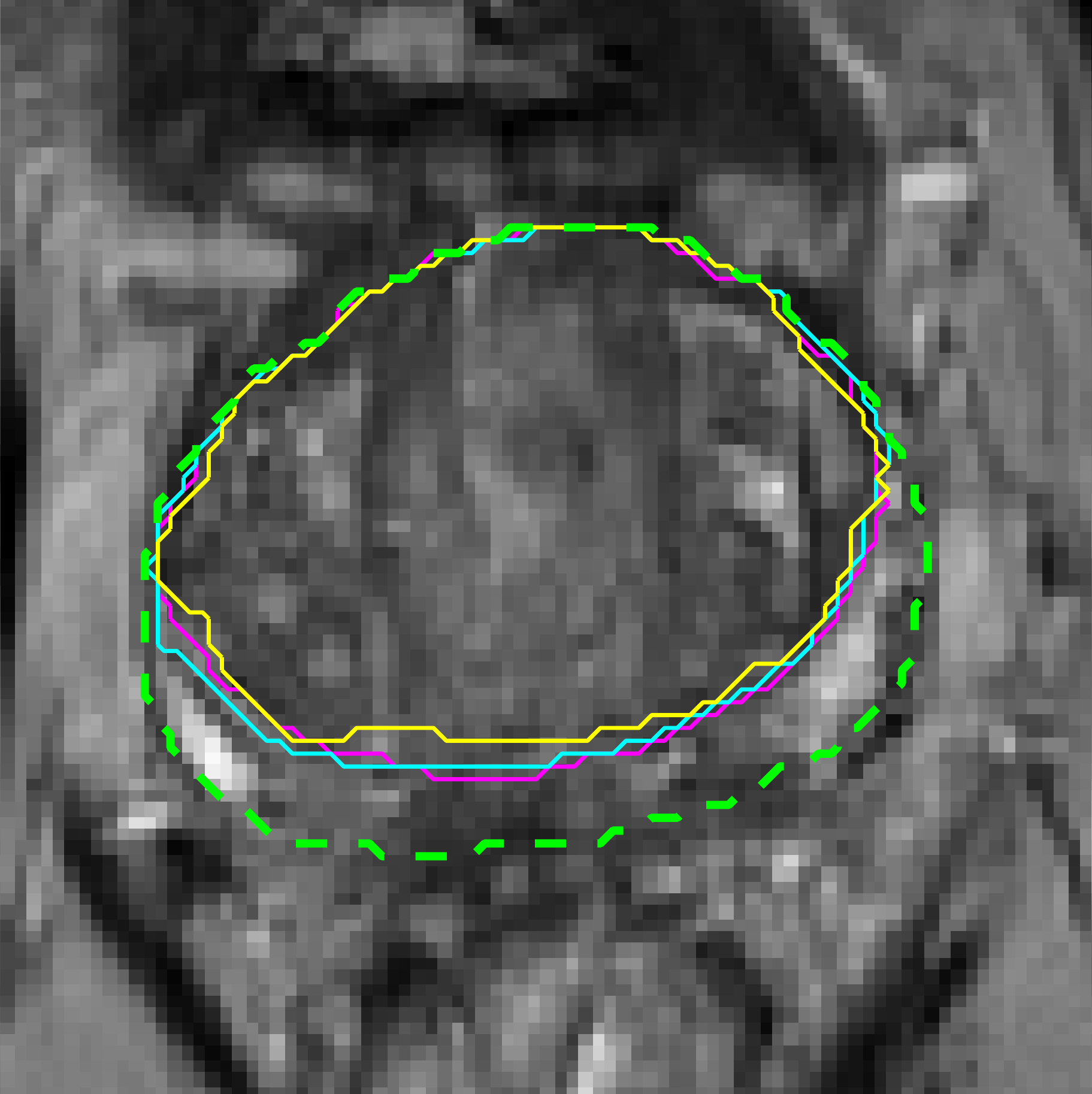}\label{fig:ResultImagesG}}\qquad
\subfloat[]{\includegraphics[height=2.5cm]{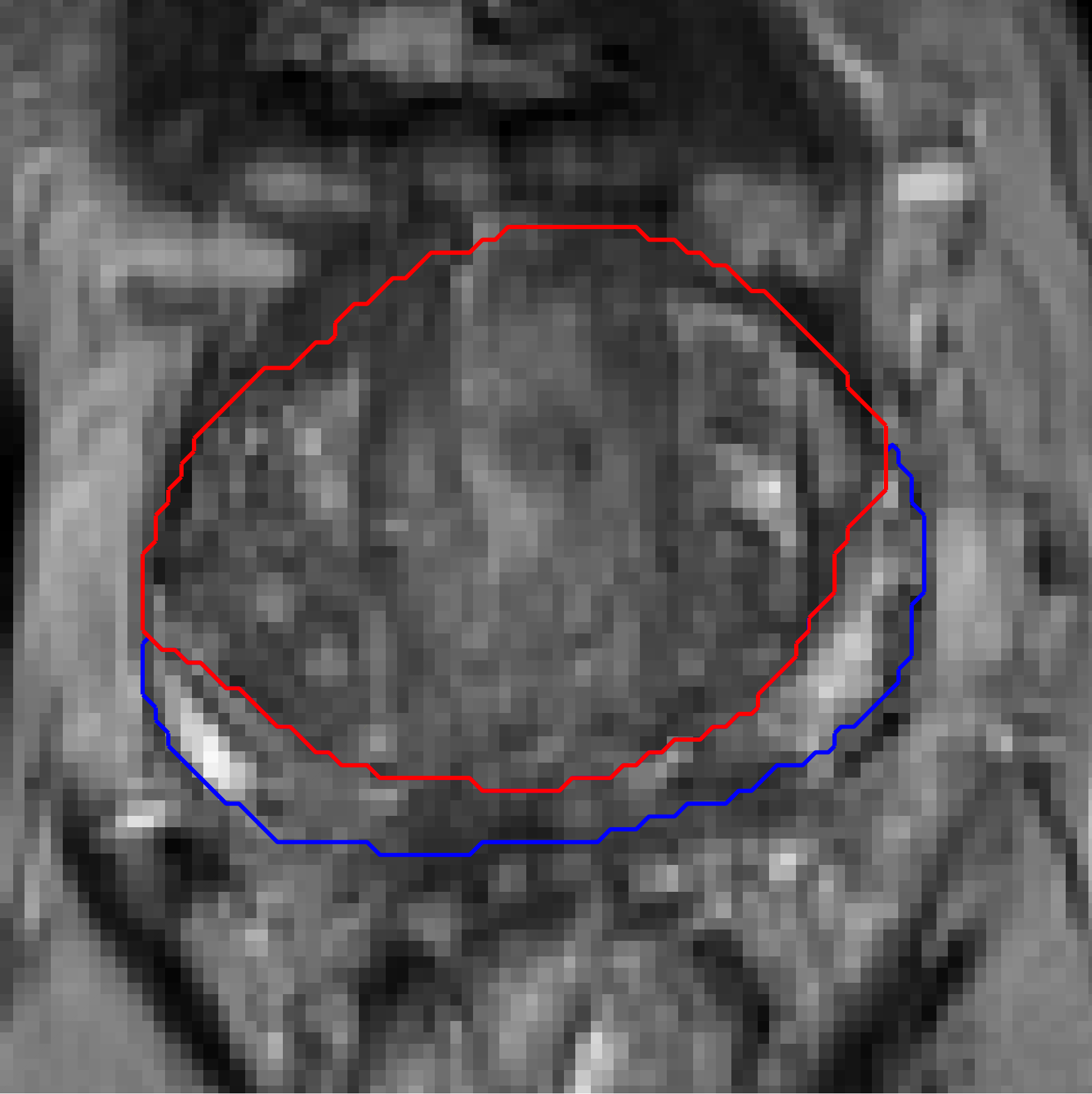}\label{fig:ResultImagesH}} \\
	\caption{Examples of prostate zonal segmentation in pre-training/fine-tuning. The first row concerns testing on dataset $\#1$, trained on: (a) dataset $\#1$; (b) dataset $\#2$; (c) mixed dataset. The second row concerns testing on dataset $\#2$, trained on: (e) dataset $\#1$; (f) dataset $\#2$; (g) mixed dataset.
    The $\mathcal{R}_{CG}$ segmentation results are represented with magenta, cyan, and yellow solid contours for SegNet, U-Net, and pix2pix, respectively.
The dashed green line denotes the $\mathcal{R}_{WG}$ boundary.
The last column (sub-figures (d) and (h)) shows the gold standard for $\mathcal{R}_{CG}$ and $\mathcal{R}_{PZ}$ with red and blue lines, respectively.
The images are zoomed with a $4\times$ factor.}
	\label{fig:ResultImages}	
\end{figure}

For a visual assessment, two examples (one for each dataset) are shown in Fig. \ref{fig:ResultImages}.
Relying on the gold standards in Figs. \ref{fig:ResultImagesD} and  \ref{fig:ResultImagesH}, it can be seen that U-Net generally achieves more accurate results compared with SegNet and pix2pix.
This finding confirms the trend revealed by the \emph{DSC} values in Table \ref{table:results}.

\section{Discussion and Conclusions}
\label{sec:Conclusions}

Our preliminary results show that CNN-based architectures can segment prostate zones on two different MRI datasets to some extent, leading to valuable clinical insights; CNNs suffer when training and testing are performed on different MRI datasets acquired by different devices and protocols, but this can be mitigated by training the CNNs on multiple datasets, even without pre-training.
Generally, considering different experimental training and testing conditions, U-Net outperforms SegNet and pix2pix thanks to its good generalization ability.
Furthermore, this study suggests that significant performance improvement \textit{via} fine-tuning may require a remarkably large dataset for pre-training.

As future developments, we plan to improve the results by refining the predicted binary masks for better smoothness and continuity, avoiding disconnected segments; furthermore, we should enhance the output delineations considering the three-dimensional spatial information among slices.
Furthermore, relying on the encouraging cross-dataset capability of U-Net, it is worth to devise and test new solutions aiming at improving the performance of the standard U-Net architecture~\cite{falk2019}.
Finally, for better cross-dataset generalization, additional prostate zonal datasets and domain adaptation using transfer learning with Generative Adversarial Networks (GANs)~\cite{goodfellow2014,han2018} and Variational Auto-Encoders (VAEs)~\cite{kingma2013} could be useful.

\section*{Acknowledgment}
This work was partially supported by the Graduate Program for Social ICT Global Creative Leaders of The University of Tokyo by JSPS.

%
%

\bibliographystyle{splncs} 
\bibliography{biblio}

\end{document}